\documentclass{article}
\usepackage{geometry}              		
\usepackage[dvipdfmx]{graphicx}
\usepackage{xcolor}				
\usepackage[T1]{fontenc}
\usepackage{mathptmx}
\usepackage{float}
\usepackage{amssymb}
\usepackage[utf8]{inputenc}
\usepackage{amsmath}
\usepackage{amsthm}
\usepackage{amsfonts}
\usepackage{bm}
\usepackage{hyperref}
\usepackage{tikz}
\usepackage{pgfplots}
\pgfplotsset{compat=1.5}
\usetikzlibrary{shapes.geometric}
\newcommand{\norm}[1]{\left\lVert#1\right\rVert}
\usepackage{algorithm}
\usepackage{algpseudocode}
\usepackage{subcaption}
\usepackage{authblk}

\DeclareMathOperator*{\argmin}{arg\,min}
\newcommand{\RomanNumeralCaps}[1]
    
\begin{document}

\title{Trajectory Estimation in Unknown Nonlinear Manifold Using Koopman Operator Theory}
\date{}

\author[1]{\small Yanran Wang}
\author[2]{\small Michael J. Banks}
\author[2]{\small Igor Mezic}
\author[1]{\small Takashi Hikihara}

\affil[1]{\footnotesize Department of Electrical Engineering, Kyoto University}
\affil[2]{\footnotesize Department of Mechanical Engineering, University of California, Santa Barbara}

\maketitle

\begin{abstract}
  Formation coordination is a critical aspect of swarm robotics, which involves coordinating the motion and behavior of a group of robots to achieve a specific objective. In formation coordination, the robots must maintain a specific spatial arrangement while in motion. In this paper, we present a leader-follower column formation coordination problem in an unknown, two-dimensional nonlinear manifold, where we redefining it as a trajectory estimation problem. Leveraging Koopman operator theory and Extended Dynamic Mode Decomposition, we estimate the measurement vectors for the follower agent and guide its nonlinear trajectories.

\end{abstract}

\section{Introduction}
Swarm Intelligence (SI) refers to the collective behavior of decentralized and self-organized systems. Individual agents in a swarm possess limited sensing abilities and operate based on simple computational rules that facilitate local interactions with each other. Nevertheless, the swarm as a whole exhibits emergent global behaviors that are not apparent in the actions of the individual agents. Bird flocking and fish schooling are examples of SI systems in nature. Researchers strive to comprehend these natural SI systems by representing them as dynamical systems \cite{flockingattanasi}, \cite{flockinggazi}, \cite{flockingmateo}, \cite{flockingsaber}, \cite{Vicsekflocking}. Inspired by natural SI systems, researchers also create artificial multi-robot systems (MRS) for a diverse range of applications \cite{applychen}, \cite{applymcguire}, \cite{applyzhang}, \cite{applyMendonca}, \cite{applyGregory}, \cite{applyshapira}.

Formation coordination is a critical issue in MRS. There are many ways to characterize and categorize existing formation coordination strategies, including leader-follower strategies and artificial potential fields, among others \cite{surveyOH}, \cite{surveyLiu}. Each strategy has its own strengths and weaknesses that must be taken into account.

In their study, Wang et al. \cite{Wang} establish a connection between Wireless Sensor Networks (WSN) and SI systems. Their research focuses on extracting environmental information by analyzing the changes in swarm formation in response to external stimuli. One of the primary challenges they encounter is maintaining formation in the presence of external environmental changes.

In this paper, we want to further address formation coordination challenge presented in \cite{Wang}. We consider the swarm is travelling in a Riemannian manifold, $\mathcal{M}$, defined by the set of solutions to a single equation, $F(x_1, \dots , x_n) = 0,$ where $F$ is a $\mathbb{C}^1$  function. We demonstrate how the formation coordination problem in swarm robotics can be reinterpreted as a trajectory estimation problem for individual agents in a leader-follower framework with unidirectional observations. In this framework, we regard each follower agent in the swarm as a nonlinear dynamical system. Employing Koopman operator theory and Extended Dynamic Mode Decomposition (EDMD), we estimate follower trajectories and offer guidance to the followers, leading to enhanced formation coordination efficiency.

Section \ref{Problem formulation} provides a mathematical definition of the formation and discusses the formation coordination problem in the leader-follower framework. In Section \ref{Proposed approach}, we present our trajectory estimation algorithm, while Section \ref{Simulations and discussions} presents the results of simulations conducted to validate the algorithm's effectiveness.

\section{Problem formulation}
\label{Problem formulation}
\subsection{Swarm formation on arbitrary Riemannian manifold}
\label{swarm formation}
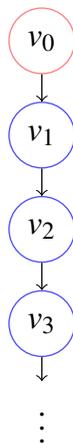
\begin{figure}[!ht]
  \centering
  \resizebox{1cm}{!}{\begin{tikzpicture}
\node[circle,
minimum width =5pt ,
minimum height =5pt ,draw=red!50] (1) at(0,0){$v_0$};
\node[circle,
minimum width =10pt ,
minimum height =10pt ,draw=blue!70] (2) at(0,-1){$v_1$};
\node[circle,
minimum width =10pt ,
minimum height =10pt ,draw=blue!70] (3) at(0,-2){$v_2$};
\node[circle,
minimum width =10pt ,
minimum height =10pt ,draw=blue!70] (4) at(0, -3){$v_3$};
\node (5) at (0, -4){$\vdots$};
\draw[->] (1) --(2);
\draw[->] (2) --(3);
\draw[->] (3) --(4);
\draw[->] (4) --(5);
\end{tikzpicture}}
  \caption{Swarm formation scheme. A directed path graph. Each vertex represents an agent in the swarm, the edges represent the inter-agent distance.}
  \label{directed_path_graph}
\end{figure}

The topology of the swarm formation follows \cite{Wang}, which is a column shape formation. Mathematically, this formation can be defined as a directed path graph $P_n = (V, E)$ consists of a set of vertices $\mathcal{V} = \{v_0,v_1, \dots, v_n\}$ and a set of edges $\mathcal{E}$ such that $\mathcal{E} \subseteq \{v_i, v_{i+1}\}$, where $i=0, 1, \dots, n-1$, as shown in Figure \ref{directed_path_graph}. Each vertex represents an individual agents in the swarm, while the edges represent the distance between any two neighbouring agents. To define the edges, let $q=(q_0,...,q_n)^T$ be the configuration of all agents, where $q_i \in \mathbb{R}$ denote the position of agent $v_i$ for all $v_i \in \mathcal{V}$. The length of edges, $dis(\mathcal{E})$, is defined to be the geodesic distance between two connected vertices over the graph $P_n(q)$.

Using the notations above, we can define the desired formation of the swarm as a path graph with the constraint
\begin{align}
 dis(\mathcal{E}_{v_i,v_{i+1}}) = d, \hspace{5mm} \forall \,\,v_i \in V, \hspace{5mm} d \in \mathbb R.
 \label{latticeconstraint}
\end{align}
We denote a formation that satisfies this constraint as the ideal formation.

\begin{figure}[!ht]
    \begin{subfigure}[b]{0.49\textwidth}
         \centering
         \resizebox{0.9\linewidth}{!}{\begin{tikzpicture}
    \node at (0.5,9) {$v_0$};
    \node at (0.5,7) {$v_1$};
    \node at (0.5,5) {$v_2$};
  \node[draw,
  fill=red!50,
  isosceles triangle,
  isosceles triangle apex angle=30,
  rotate=0,inner sep =5pt] at (1,9) (nodeA1){};
  \node[draw,
  fill=red!50,
  isosceles triangle,
  isosceles triangle apex angle=30,
  rotate=0,inner sep =5pt] at (6,9) (nodeA2){};
  \node[draw,
  fill=red!50,
  isosceles triangle,
  isosceles triangle apex angle=30,
  rotate=0,inner sep =5pt] at (12,9) (nodeA3){};

\node[draw,
fill=blue!70,
isosceles triangle,
isosceles triangle apex angle=30,
rotate=0,inner sep =5pt] at (1,7) (nodeB1){};
\node[draw,
fill=blue!70,
isosceles triangle,
isosceles triangle apex angle=30,
rotate=0,inner sep =5pt] at (6,7) (nodeB2){};
\node[draw,
fill=blue!70,
isosceles triangle,
isosceles triangle apex angle=30,
rotate=0,inner sep =5pt] at (12,7) (nodeB3){};

\node[draw,
fill=blue!50,
isosceles triangle,
isosceles triangle apex angle=30,
rotate=0,inner sep =5pt] at (1,5) (nodeC1){};
\node[draw,
fill=blue!50,
isosceles triangle,
isosceles triangle apex angle=30,
rotate=0,inner sep =5pt] at (6,5) (nodeC2){};
\node[draw,
fill=blue!50,
isosceles triangle,
isosceles triangle apex angle=30,
rotate=0,inner sep =5pt] at (12,5) (nodeC3){};

\draw[line width=0.5mm, red!50] (nodeA1) -- (nodeA2) -- (13,9);
\draw[line width=0.5mm, blue!70] (nodeB1) -- (nodeB2) -- (13,7);
\draw[line width=0.5mm, blue!50] (nodeC1) -- (nodeC2) -- (13,5);

\node at (nodeA1) [circle,fill,inner sep=1.5pt]{};
\node at (nodeB1) [circle,fill,inner sep=1.5pt]{};
\node at (nodeC1) [circle,fill,inner sep=1.5pt]{};
\node at (nodeA2) [circle,fill,inner sep=1.5pt]{};
\node at (nodeB2) [circle,fill,inner sep=1.5pt]{};
\node at (nodeC2) [circle,fill,inner sep=1.5pt]{};
\node at (nodeA3) [circle,fill,inner sep=1.5pt]{};
\node at (nodeB3) [circle,fill,inner sep=1.5pt]{};
\node at (nodeC3) [circle,fill,inner sep=1.5pt]{};

\draw[dashed, line width=0.2mm, black] (nodeA1) -- (nodeB1);
\draw[dashed, line width=0.2mm, black] (nodeA2) -- (nodeB2);
\draw[dashed, line width=0.2mm, black] (nodeA3) -- (nodeB3);
\draw[dashed, line width=0.2mm, black] (nodeB1) -- (nodeC1);
\draw[dashed, line width=0.2mm, black] (nodeB2) -- (nodeC2);
\draw[dashed, line width=0.2mm, black] (nodeB3) -- (nodeC3);

  \end{tikzpicture}}
         \caption{Swarm formation on flat plane.}
         \label{formation_flat}
    \end{subfigure}
    \begin{subfigure}[b]{0.49\textwidth}
         \centering
         \resizebox{0.9\linewidth}{!}{\begin{tikzpicture}
    \node at (0.8,9) {$v_0$};
    \node at (0.8,7) {$v_1$};
    \node at (0.8,5) {$v_2$};
  \node[draw,
  fill=red!50,
  isosceles triangle,
  isosceles triangle apex angle=30,
  rotate=0,inner sep =5pt] at (2,9) (nodeA1){};
  \node[draw,
  fill=red!50,
  isosceles triangle,
  isosceles triangle apex angle=30,
  rotate=-6,inner sep =5pt] at (7.5,8.8) (nodeA2){};
  \node[draw,
  fill=red!50,
  isosceles triangle,
  isosceles triangle apex angle=30,
  rotate=-15,inner sep =5pt] at (13,7.8) (nodeA3){};

\node[draw,
fill=blue!70,
isosceles triangle,
isosceles triangle apex angle=30,
rotate=-15,inner sep =5pt] at (1.8,7) (nodeB1){};
\node[draw,
fill=blue!70,
isosceles triangle,
isosceles triangle apex angle=30,
rotate=-5,inner sep =5pt] at (7.2,6.15) (nodeB2){};
\node[draw,
fill=blue!70,
isosceles triangle,
isosceles triangle apex angle=30,
rotate=-6,inner sep =5pt] at (12.3,5.7) (nodeB3){};

\node[draw,
fill=blue!50,
isosceles triangle,
isosceles triangle apex angle=30,
rotate=-25,inner sep =5pt] at (1.9,5) (nodeC1){};
\node[draw,
fill=blue!50,
isosceles triangle,
isosceles triangle apex angle=30,
rotate=1,inner sep =5pt] at (7.6,3.3) (nodeC2){};
\node[draw,
fill=blue!50,
isosceles triangle,
isosceles triangle apex angle=30,
rotate=5,inner sep =5pt] at (12.5,3.4) (nodeC3){};

\draw[line width=0.5mm, red!50] (nodeA1) .. controls (4, 9) and (10, 9) .. (14, 7.5);
\draw[line width=0.5mm, blue!70] (nodeB1) .. controls (5, 6.2) and (10, 6) .. (14, 5.5);
\draw[line width=0.5mm, blue!50] (nodeC1) .. controls (6, 3) and (9, 3.2) .. (14, 3.5);

\node at (nodeA1) [circle,fill,inner sep=1.5pt]{};
\node at (nodeB1) [circle,fill,inner sep=1.5pt]{};
\node at (nodeC1) [circle,fill,inner sep=1.5pt]{};
\node at (nodeA2) [circle,fill,inner sep=1.5pt]{};
\node at (nodeB2) [circle,fill,inner sep=1.5pt]{};
\node at (nodeC2) [circle,fill,inner sep=1.5pt]{};
\node at (nodeA3) [circle,fill,inner sep=1.5pt]{};
\node at (nodeB3) [circle,fill,inner sep=1.5pt]{};
\node at (nodeC3) [circle,fill,inner sep=1.5pt]{};

\draw[densely dotted, line width=0.2mm, black] (nodeA1) .. controls (1.7, 7.8) .. (nodeB1);
\draw[densely dotted, line width=0.2mm, black] (nodeB1) .. controls (1.75, 5.8) .. (nodeC1);
\draw[densely dotted, line width=0.2mm, black] (nodeA2) .. controls (7.1, 7.1) .. (nodeB2);
\draw[densely dotted, line width=0.2mm, black] (nodeB2) .. controls (7.2, 4.8) .. (nodeC2);
\draw[densely dotted, line width=0.2mm, black] (nodeA3) .. controls (12.3, 6.8) .. (nodeB3);
\draw[densely dotted, line width=0.2mm, black] (nodeB3) .. controls (12.2, 4.5) .. (nodeC3);

\draw[dashed, line width=0.2mm, black] (nodeA1) -- (nodeB1);
\draw[dashed, line width=0.2mm, black] (nodeA2) -- (nodeB2);
\draw[dashed, line width=0.2mm, black] (nodeA3) -- (nodeB3);
\draw[dashed, line width=0.2mm, black] (nodeB1) -- (nodeC1);
\draw[dashed, line width=0.2mm, black] (nodeB2) -- (nodeC2);
\draw[dashed, line width=0.2mm, black] (nodeB3) -- (nodeC3);

  \end{tikzpicture}}
         \caption{Swarm formation on arbitrary Riemannian manifold.}
         \label{formation_curve}
    \end{subfigure}
    \caption{Swarm column shape formation moving on flat plane and unknown two dimensional Riemannian surface. The dotted line represents the inter-agent distance in Riemannian metric which have identical distance measured in particular Riemannian metric. The dashed line represents the inter-agent distance measured in Euclidean metric. The two types of measurements yield the same result when dealing with a flat plane, whereas the measurements differs for Riemannian surface.}
    \label{formation_on_surface}
\end{figure}
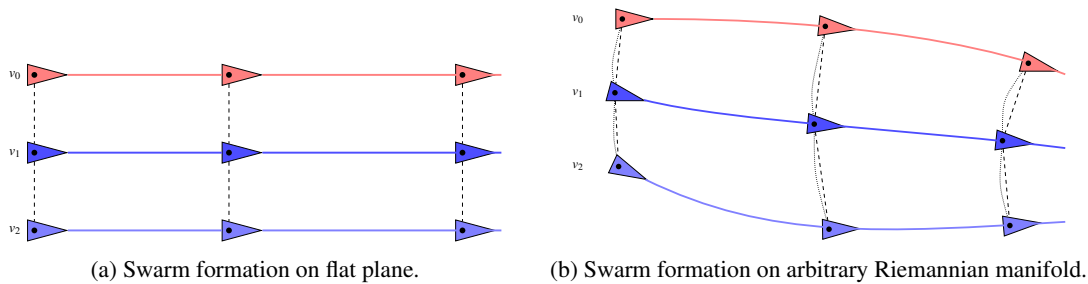

On a flat plane, the ideal formation can be represented as a graph with all its vertices and edges lying on a single straight line, with identical inter-agent distances. From the perspective of an individual agent, this distance is measured along a line that is orthogonal to its velocity. By knowing the leader's velocity, the follower can maintain its formation. However, this approach cannot be applied if the formation moves on an arbitrary two-dimensional Riemannian manifold. Figure \ref{formation_on_surface} provides a visual comparison between the two scenarios.

To express the defined ideal formation on an arbitrary two-dimensional Riemannian manifold $\mathcal{M}$ in mathematical terms, we begin by assuming that agent $v_0$ has a defined trajectory $l(\mathbf{x},t)$ on $\mathcal{M}$. We then define the edge to be a parallel vector field orthogonal to $l(\mathbf{x},t)$ in the metric $g$ of $\mathcal{M}$, and the trajectories of the remaining agents $v_1, v_2, \ldots$, are the integral curves of the geodesic deviation vector field. A detailed discussion of the theoretical basis and formulation is provided in Appendix \ref{Swarm formation construction} \cite{Wang}.

\subsection{Formation coordination to trajectory estimation}

The objective of the formation coordination is for the swarm to maintain the formation constraint, i.e., identical inter-agent distance, as it moves along an unknown two-dimensional Riemannian manifold.

In pursuit of a decentralized approach, we propose a swarm formation that relies on relative distance-based formation coordination in a leader-follower framework. The agents are divided into two groups: leaders and followers. The swarm consists of one primary leader that provides trajectory guidance and motion stimulus, and a group of followers that follow the primary leader while adhering to the inter-agent distance constraint described above. As the swarm takes a column shape formation, each follower also acts as a secondary leader for its neighboring followers. The agents in the swarm are assumed to be slow-moving and have access only to local and relative information (relative distance and relative orientation).

There are two distinct types of information that can be conveyed within a swarm. Firstly, followers are assumed to possess the ability to sense the relative Euclidean distance measurements related to their leaders, based on their own local coordinate system. Secondly, primary and secondary leaders offer Riemannian distance measurements through a physical extension that is perpendicular to their path in a Riemannian metric, ensuring a consistent and pre-set Riemannian distance is maintained. We refer to the termination point of the extension as the follower's ideal position since it satisfies the swarm formation distance constraint described in Equation \ref{latticeconstraint}. The followers have the ability to passively observe this ideal position provided by the leader's extension, utilizing various means such as visual cameras. 

\begin{figure}[!ht]
    \begin{center}
      \includegraphics[width=0.48\textwidth]{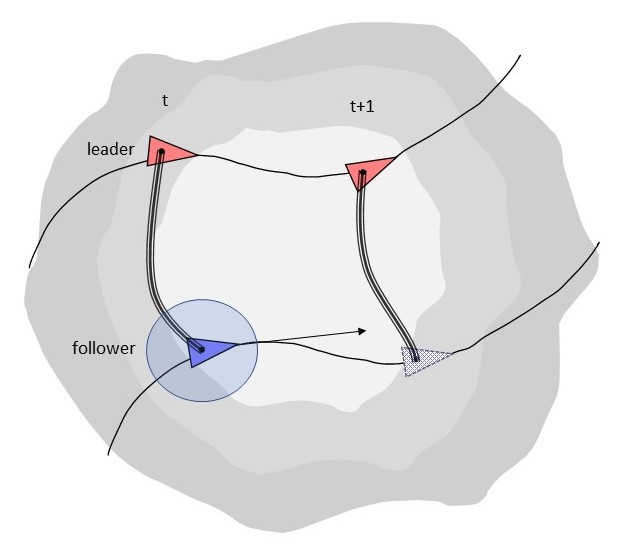}
    \end{center}
    \caption{Illustration of formation. The leader is denoted by the red triangle, while the follower is represented in blue. The extension, depicted by the tripled line, measures a fixed distance orthogonal to the leader's current heading in the metric of the unknown Riemannian manifold. The follower's sensing range is illustrated by the blue circle. The transparent blue triangle indicates the follower's ideal position at timestep $t+1$.}
    \label{illustration_follwermove}
\end{figure}

The termination point of the leader extension may exceed the sensing range of followers, depending on the communication frequency between leaders and followers, as shown in Figure. \ref{illustration_follwermove}. To ensure efficient movement of the swarm while maintaining the desired formation, followers need the ability to anticipate and estimate the ideal position provided by their leaders' extensions using available data. Essentially, the problem of formation coordination within the swarm can be seen as an issue of trajectory estimation for individual followers. This estimation allows them to adjust their own movements accordingly and stay in the formation defined by the constraint.

By reframing the formation coordination problem as a trajectory estimation challenge, the goal of preserving the ideal formation defined by the constraint equation \ref{latticeconstraint} is reshaped into the task of utilizing historical data to provide a one-timestep estimation of the follower's trajectory when it adheres to the ideal formation.

\section{Proposed approach}
\label{Proposed approach}
The trajectory of the individual follower can be considered as a nonlinear dynamic system of the form \cite{strogatz}
\begin{align}
      \frac{d}{dt} \mathbf{x}(t)= \mathcal{F} (\mathbf{x}(t))
\label{follower_dynamics}
\end{align}
where $\mathbf{x}(t) \in \mathcal{M}$ is the follower's position, and $\mathbf{F}$ is a vector field depending on the manifold. In practice we consider the equivalent discrete-time dynamic system
\begin{align}
  \mathbf{x}_{t+1} = \mathcal{F}(\mathbf{x}_t),
\end{align}
where $\mathbf{x}_t$ is the sampling of the agent trajectory in (\ref{follower_dynamics}) discretely in timesteps $\Delta t$. 

The follower needs to rely on available past data to make a one-timestep prediction on its trajectory. The challenge lies in the fact that we lack information about the Riemannian manifold, resulting in an unknown nonlinear vector field represented by $\mathcal{F}$. In order to analyze the nonlinear dynamics using a linear technique, we employ the Koopman operator theory and EDMD.

\subsection{Koopman operator theory and Extended Dynamic Mode Decomposition}
Koopman operator theory grants the possibility of representing a nonlinear dynamical system in terms of an infinite-dimensional linear operator acting on a Hilbert space of measurement functions of the state of the system \cite{Koopmanbook}, \cite{Koopmanpaper}.

Consider a nonlinear discrete-time dynamical system,
\begin{align}
  \mathbf{x}_{t+1} = \mathcal{F}(\mathbf{x}_t)
\end{align}
where $\mathbf{x}_t \in \mathcal{M}$ is the state at timestep $t$. 
Also consider a real-valued measurement function $g : \mathcal{M} \rightarrow \mathbb{R}$, known as observables, which belongs to the infinite-dimensional
Hilbert space. The Koopman operator $\mathcal{K}$ is an infinite-dimensional linear operator that acts on the scalar observable $g: \mathcal{M} \longrightarrow \mathbb{R}$ as
\begin{align}
  \mathcal{K}g(\mathbf{x}_t) =  g \circ (\mathcal{F}(\mathbf{x}_t)).
\end{align}

Despite the infinite-dimensional nature of the Koopman operator, its linear characteristics allow us to perform eigendecompositions and effectively capture the linear evolution of key measurement functions within the dynamics \cite{Mezic_decomposition}. However, obtaining the Koopman modes analytically can be exceedingly difficult, especially in cases where the governing equation of the dynamical system is completely unknown. Therefore, a data-driven method is required. In our study, we have chosen to utilize the Extended Dynamic Mode Decomposition (EDMD) approach \cite{EDMD_mezic,EDMD_rowley,Mezic_koopmanMPC}. EDMD is a data-driven method that can approximate the Koopman operator without requiring explicit knowledge of the system's governing equations.

Consider a data set of snapshot pairs $\{(\mathbf{x}_t,\mathbf{y}_t)\}^m_{t=1}$, where $\mathbf{x}_t \in \mathcal{M}$ and $\mathbf{y}_t \in \mathcal{M}$ are snapshots of the dynamic system with $\mathbf{y}_t = \mathcal{F}(\mathbf{x}_t)$. Also consider a dictionary of observables, $\mathcal{D} = \{ \psi_1, \psi_2, \dots, \psi_k \}$, where $\psi_i: \mathcal{M} \rightarrow \mathbb{R}$, then we have a vector valued function $\Psi: \mathcal{M} \rightarrow \mathbb{R}^{k}$ with
\begin{align}
  \Psi(\mathbf{x})=\begin{bmatrix}
    \psi_1(\mathbf{x}) \\
    \psi_2(\mathbf{x}) \\
   \vdots          \\
   \psi_k(\mathbf{x}) \\
  \end{bmatrix}.
 \end{align}

Now consider the span $U(\Psi) = \text{span}\{\psi_1, \dots, \psi_k\} = \{a^T\Psi:a \in \mathbb{C}^k\}.$ Then, for a function $g = a^T \Psi \in U(\Psi)$, we have 
\begin{align}
  \mathcal{K}g = a^T\mathcal{K}\Psi = a^T \Psi \circ \mathcal{F}. 
  \label{dictionary_koopman}
\end{align}
A finite dimensional representation of the Koopman operator $\mathcal{K}$ is the matrix $\mathbf{K} \in \mathbb{R}^{k \times k}$. Then, for Equation (\ref{dictionary_koopman}) to hold for all $a$, we have
\begin{align}
  \mathbf{K} \Psi = \Psi \circ \mathcal{F}.
\end{align}
To compute $\mathbf{K}$, we slove the minimization problem, 
\begin{align}
  \mathbf{K} = \argmin_{\tilde{\mathbf{K}} \in \mathbb{R}^{k \times k}} \sum^{m}_{t = 1} \norm{\Psi(\mathbf{y}_t) - \tilde{\mathbf{K}}\Psi (\mathbf{x}_t)}^2
\end{align}

\subsection{Numerical Algorithm}
In summary, EDMD facilitates a one-timestep estimation of the measurements of system states. This estimation is achieved by approximating the finite-dimensional Koopman operator that advances the dyanmics of the follower. 

Consider a set of data
\begin{align}
 \mathbf{X} & =\begin{bmatrix}
  \vert        & \vert        &       & \vert            \\
  \mathbf{x}_1 & \mathbf{x}_2 & \dots & \mathbf{x}_{m} \\
  \vert        & \vert        &       & \vert            \\
 \end{bmatrix},
 \\
 \mathbf{Y} & =\begin{bmatrix}
  \vert        & \vert        &       & \vert          \\
  \mathbf{y}_1 & \mathbf{y}_2 & \dots & \mathbf{y}_{m} \\
  \vert        & \vert        &       & \vert          \\
 \end{bmatrix}.
\label{data_matrix}
\end{align}
satisfies the relation $\mathbf{y}_i = \mathcal{F}(\mathbf{x}_i)$.

By passing the data $\mathbf{X}, \mathbf{Y}$ through a dictionary of observables, we obtain its lifted representation, denoted as $\mathbf{X}_{\text{lifted}}$ and $\mathbf{Y}_{\text{lifted}}$ with 
\begin{align}
  \mathbf{X}_{\text{lifted}} & =\begin{bmatrix}
   \vert        & \vert        &       & \vert            \\
   \Psi(\mathbf{x}_1) & \Psi(\mathbf{x}_2) & \dots & \Psi(\mathbf{x}_{m}) \\
   \vert        & \vert        &       & \vert            \\
  \end{bmatrix},
  \\
  \mathbf{Y}_{\text{lifted}} & =\begin{bmatrix}
   \vert        & \vert        &       & \vert          \\
   \Psi(\mathbf{y}_1) & \Psi(\mathbf{y}_2) & \dots & \Psi(\mathbf{y}_{m}) \\
   \vert        & \vert        &       & \vert          \\
  \end{bmatrix}.
 \label{data_matrix_edmd}
 \end{align}

Given $\mathbf{X}_{\text{lifted}},\mathbf{Y}_{\text{lifted}}$ in \ref{data_matrix_edmd}, the matrix $\mathbf{K} \in \mathbb{R}^{k \times k}$ that approximates the Koopman operator of the nonlinear dynamical system are obtained by solving the least-sqaure problem 
\begin{align}
  \mathbf{K} = \mathop{\arg\min}_{\tilde{\mathbf{K}}} \norm{\mathbf{Y}_{\text{lifted}}-\mathbf{\tilde{\mathbf{K}} \mathbf{X}_{\text{lifted}}}}_F,
 \end{align}
 where $\norm{\cdot}_F$ is the Frobenius norm.
 The unique solution to least-square problem is given by
 \begin{align}
  \mathbf{K}=\mathbf{\mathbf{Y}_{\text{lifted}}\mathbf{X}_{\text{lifted}}}^{\dagger},
  \label{matrix a}
 \end{align}
 where $\mathbf{X}_{\text{lifted}}^{\dagger}$ denotes the Moore-Penrose pseudoinverse of $\mathbf{X}_{\text{lifted}}$.

 In our specific scenario, we lack access to the system state data $\mathbf{X}$ and $\mathbf{Y}$. Instead, we utilize measurement data from the lifted space $\mathbf{X_{\text{lifted}}}$ and $\mathbf{Y_{\text{lifted}}}$. Notably, even if the observable functions $\Psi$ may not be explicitly known, the prediction algorithm remains effective. Additionally, new measurement data is continuously introduced at each time step, which necessitates the implementation of a data-streaming adaptation of the EDMD algorithm. 

The stepwise process of this algorithm is detailed in the provided pseudo-code example, Algorithm \ref{alg:edmd}. The prediction procedure unfolds as follows: leveraging available measurement data at timestep $m$, the algorithm estimates a finite-dimensional approximation of the Koopman operator, denoted as $\mathbf{K}$. Subsequently, a one-timestep prediction is generated based on all measurement data up to timestep $m$ and the approximated Koopman operator $\mathbf{K}$ at the same timestep. This predicted new measurement data, denoted as $\mathbf{y}^\ast_{m+1}$, undergoes correction by the newly acquired actual measurement data $\mathbf{y}_{m+1}$. The iterative nature of this process involves incorporating the newly acquired measurement data into the historical dataset. This continuous cycle allows the algorithm to adapt and refine predictions as new information becomes available. Please take note that $\mathbf{x}_i$ and $\mathbf{y}_i$ in Algorithm \ref{alg:edmd} denote data from the lifted space, and their subscripts are omitted for simplicity.

\begin{algorithm}[h]
  \caption{Data-streaming EDMD}\label{alg:edmd}
  \begin{algorithmic}
  \Require Data set of snapshot pairs ${\{\mathbf{x}_t, \mathbf{y}_t\}}^m_{t=1}$, with $\mathbf{y}_t = \mathbf{x}_{t+1}$.
  \Require Iterations for prediction, $M$
  \State $\mathbf{X}_m \gets \mathbf{x}_1, ... , \mathbf{x}_m$
  \State $\mathbf{Y}_m \gets \mathbf{y}_1, ... , \mathbf{y}_m$
  \While{$m < M$}
  \State $\mathbf{K} \gets \mathbf{Y}_m \mathbf{X}_m^{\dagger}$
  \State $\mathbf{y}_{m+1}^\ast \gets \mathbf{K}\mathbf{y}_{m}$
  \State $\mathbf{X}_{m+1} \gets [\mathbf{X}_m, \mathbf{y}_m]$
  \State $\mathbf{Y}_{m+1} \gets [\mathbf{Y}_m, \mathbf{y}_{m+1}]$
  \State $m \gets m + 1$
  \EndWhile
  \end{algorithmic}
  \end{algorithm}

\section{Simulations and discussions}
\label{Simulations and discussions}
Since we are dealing with an unknown manifold, it is crucial for the Data-streaming EDMD algorithm to be effective across any arbitrary manifold. Despite there being infinitely many unique manifolds, we can categorize them into three types: those with positive curvature everywhere (manifold type I), those with negative curvature everywhere (manifold type II), and those with a combination of positive and negative curvature (manifold type III).

In this section, we aim to demonstrate the algorithm's effectiveness in estimating the follower's trajectory. Additionally, we will provide supportive evidence to validate that the dictionary we have chosen for the Data-streaming EDMD algorithm is the most suitable for maintaining the defined formation of a swarm in an unknown nonlinear manifold.

\subsection{Effectiveness}
\label{effectiveness}
\begin{figure*}[!h]
    \begin{center}
      \includegraphics[width=16cm]{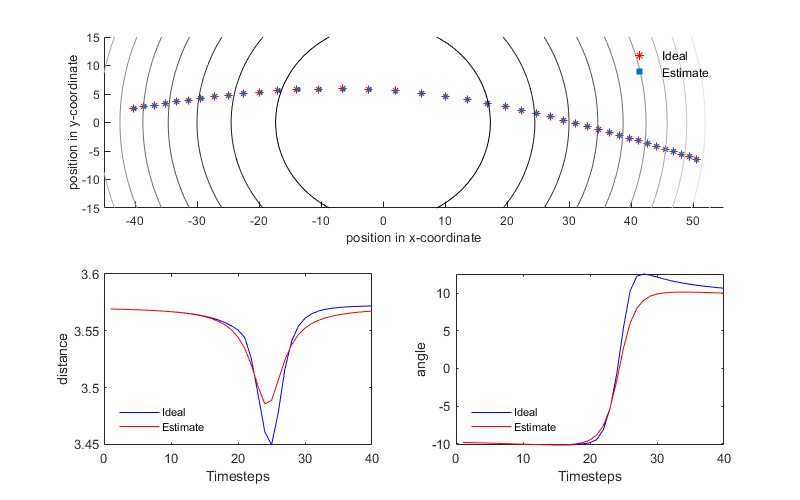}
    \end{center}
    \caption{Follower trajectory estimation by data-streaming EDMD algorithm in type I manifold. The ideal trajectory points are in red, while estimated trajectory points in blue. Manifold show as the contour lines. The estimation and ideal measurements data (distance and angle) comparison for each timesteps are shown in red and blue, respectively.}
    \label{graph_effective_el}
\end{figure*}

To showcase the efficacy of algorithm \ref{alg:edmd}, we present an illustrative example, Figure \ref{graph_effective_el}. The dictionary of the data-streaming EDMD algorithm is selected to consist of the relative position between follower and leader, measured by two-dimensional Euclidean distance and angle in the follower's coordinate system. 

In figure \ref{graph_effective_el}, for the parameters influencing the follower trajectory, we selected the manifold as an elliptic paraboloid with the function $z = \frac{1}{30} (x^2+y^2)$ (Type I manifold); the leader trajectory was defined as a geodesic curve of the manifold, starting with initial conditions $[x, y, \dot{x}, \dot{y}] = [50, -10, -\cos(\frac{1}{36}\pi), \sin(\frac{1}{36}\pi)]$; the follower maintained a distance of 3.57 units from the leader (extension length), and communication between the leader and the follower occurred at a frequency of 1.28 units of time. For the data-streaming EDMD algorithm, we used initial data pair matrices with 4 timesteps of data points, and the algorithm made estimations for a total of 36 timesteps.

We clearly observe that the algorithm excels in estimating the relative position and angle between the follower and the leader, even when both values exhibit nonlinearity over time. 

 To validate its performance across different scenarios, we simulate all three cases of manifold types (Figure \ref{graph_effective_el}, Figure \ref{graph_effective_hy}, \ref{graph_effective_trig}, respectively). The algorithm consistently demonstrates good performance across these different types of manifolds, as evidenced in the simulations provided in Appendix \ref{Illustrations of algorithm effectiveness}. 

\begin{figure}
  \begin{center}
    \includegraphics[width=8cm]{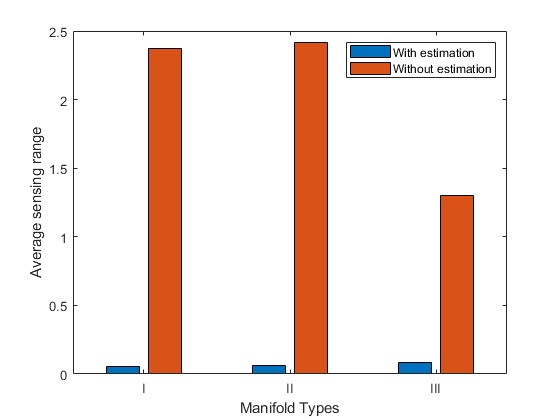}
  \end{center}
  \caption{Comparison of averaged follower required sensing range. Blue indicates the usage of the data-streaming EDMD algorithm, while red does not involve any estimation.}
  \label{eff_sensing_range_bar}
\end{figure}

 In Figure \ref{eff_sensing_range_bar}, we showcase the algorithm's estimation effectiveness by comparing the timesteps averaged sensing range needed for the follower to reach the ideal position (the terminal point of the leader's extension). Specifically, we evaluate the follower's trajectory with and without employing the data-streaming EDMD algorithm for estimation. The sensing range is computed as the 2-dimensional Euclidean distance between the follower's current position and the ideal position. When the data-streaming EDMD algorithm is applied, the current position refers to the estimated position; otherwise, it corresponds to the follower's position from the previous timestep, as there is no trajectory guidance. Remarkably, the algorithm significantly reduces the required sensing range for the follower to attain the ideal position across different manifold types provided by the leader.

\subsection{Dictionary for data-streaming EDMD algorithm}
The data-streaming EDMD algorithm offers the flexibility to select various measurement functions of the state as dictionaries. This choice significantly impacts the accuracy of the data-streaming EDMD algorithm in approximating the Koopman operator. When a well-suited dictionary is chosen, the data-streaming EDMD algorithm can outperform the conventional DMD algorithm, which utilizes system states in approximating the Koopman operator.

To align with the decentralized approach, two dictionaries are carefully selected: the follower's velocity and the leader-follower relative position (distance and angle) measured in two-dimensional Euclidean metric in follower's coordinate system. Notably, both of these dictionaries do not require global information and can be acquired by the follower through the use of on-board sensors.

\begin{figure}[!ht]
  \centering
  \begin{subfigure}[b]{0.49\textwidth}
      \centering
      \includegraphics[width=\textwidth]{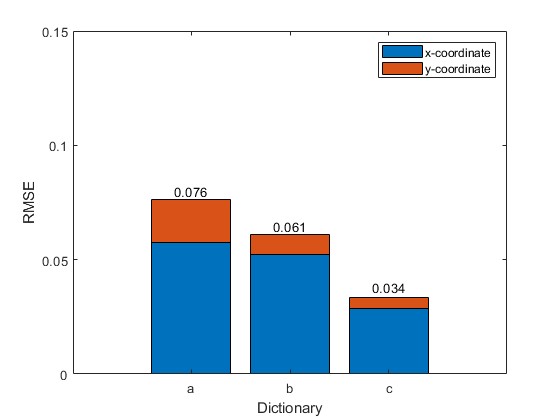}
      \caption{}
  \end{subfigure}
  \hfill
  \begin{subfigure}[b]{0.49\textwidth}
      \centering
      \includegraphics[width=\textwidth]{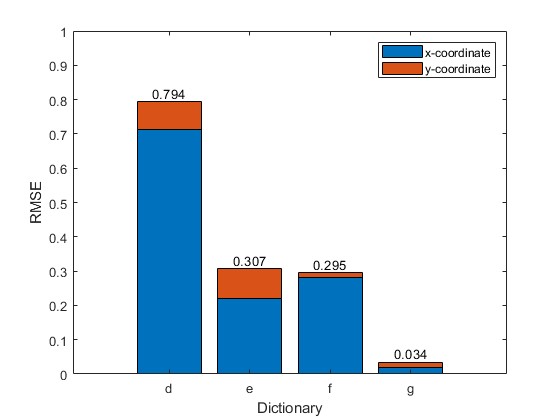}
      \caption{}
  \end{subfigure}
  \hfill
  \caption{RMSE for various dictionary choices used for data-streaming EDMD algorithm using manifold type I parameters.}
  \label{rmse_diction_el}
\end{figure}

Simulations have been conducted to compare how different dictionaries affect the accuracy of the data-streaming EDMD algorithm's estimation of ideal positions, as depicted in Figure \ref{rmse_diction_el}. The specific dictionary choices are detailed in Table \ref{dictionary_para}.

\begin{table*}
  \centering
   \begin{tabular}{||c c c c c c c||} 
   \hline
   \small Dictionary & \small follower &\small  leader-follower & \small follower 2D & \small leader 2D & \small follower 3D & \small leader 3D \\ 
   & \small velocity &\small  relative position & \small position&\small position&\small position &\small position \\
   \hline\hline
   a & \checkmark & &&&& \\ 
   b & &\checkmark & &&& \\
   c & \checkmark &\checkmark &&&& \\
   d &  & &\checkmark&&& \\
   e &  & &&&\checkmark& \\ 
   f &  & &\checkmark&\checkmark&& \\
   g &  & &&&\checkmark& \checkmark\\[0.5ex] 
   \hline
   \end{tabular}
\caption{Specific dictionary choices for EDMD algorithm.}
   \label{dictionary_para}
  \end{table*}

In Figure \ref{rmse_diction_el}.(a), for dictionary choice `a', the data-streaming EDMD algorithm estimate the one-timestep follower's velocity; while for dictionary choices `b' and `c', the algorithm estimates the one-timestep leader-follower relative position. Once the estimation is made, the follower utilizes the estimated information to travel in the manifold. If the estimated information is a velocity, the follower will travel with that velocity for 1 unit of time. On the other hand, if the estiamted information is a relative position, the follower will move to the position with the predicted value. Subsequently, the total timestep averaged Root Mean Squared Error (RMSE) is computed by comparing the estimated position with the ideal position. 

As a comparison, in Figure \ref{rmse_diction_el}.(b), we assume that the follower has access to its absolute position (system states). For dictionary choices `d' to `g', the algorithm (DMD) directly predicts the one-timestep follower's two-dimensional absolute position. The root mean square error (RMSE) is then computed by comparing this estimated position with the ideal position.

As depicted in Figure \ref{rmse_diction_el}, utilizing local information such as relative position and follower velocity yields a more accurate approximation of the Koopman operator compared to relying on system states, such as the follower's absolute position. Consequently, this approach leads to better estimation for the follower's trajectory.

The same comparisons have also been done for manifold types II and III, all of which have consistently shown that using local information to estimate follower trajectories by the data-streaming EDMD algorithm is better or on par with using system states directly. For results of manifold type II and III , please refer to Appendix \ref{dictionary for data-streaming EDMD algorithm}.

\subsection{Consideration of practical cases}

While the inclusion of self-velocity as one of the dictionaries for the data-streaming EDMD algorithm prediction, as demonstrated in Figure \ref{rmse_diction_el}, dictionary choice `c', can potentially lead to improved estimation accuracy, it's important to acknowledge that obtaining accurate follower velocity data becomes challenging in practical scenarios. This is particularly evident when considering real-world applications like formation coordination of mobile robots travel on nonlinear manifolds.

In practical scenarios, the follower faces two difficulties when accessing its accurate velocity data. The first difficulty arises from the fact that, in using both the leader-follower relative position and the follower's velocity as dictionaries for the data-streaming EDMD algorithm to predict the relative position at timestep $t+1$, we require access to both the relative position and follower's velocity data up to timestep $t$. However, obtaining the ideal follower's velocity at timestep $t$ is not feasible.

To overcome this challenge, two methods can be employed. Firstly, by assuming that changes in the follower's velocity between each timestep are insignificant, we can approximate the velocity at timestep $t$ to be equal to that of timestep $t-1$. Secondly, given that the follower's velocity can serve as one of the inputs for the data-streaming EDMD algorithm, we could execute the algorithm using only the follower's velocity data up to timestep $t-1$ to estimate the velocity at timestep $t$. By employing either of these methods, we can estimate the follower's velocity data at timestep $t$ and subsequently proceed with our data-streaming EDMD algorithm to estimate follower trajectories, utilizing both the leader-follower relative positions and the follower's velocity as inputs as dictionaries.

\begin{figure}[!ht]
  \centering
      \includegraphics[width=7cm]{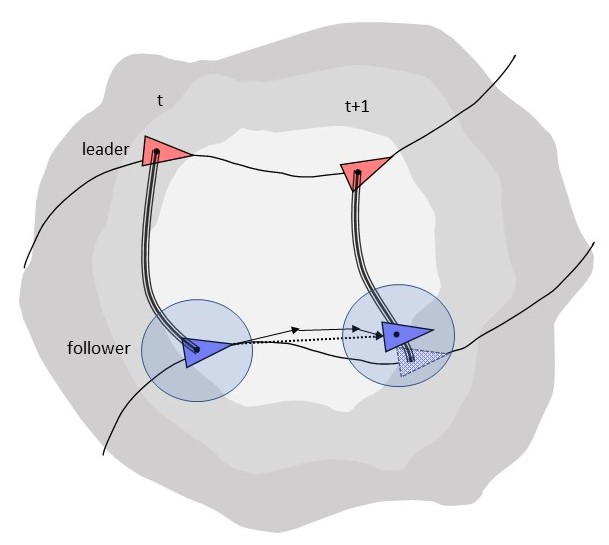}
  \caption{Illustration of follower movement in real-world cases.}
  \label{real_case_illu}
\end{figure}

The second difficulty is illustrated in Figure \ref{real_case_illu}. To reach the estimated position, the follower must iterate by comparing the current relative position with the relative position estimated using the data-streaming EDMD algorithm. However, the theoretical velocity needed to travel to the estimated position does not equal the summation of the sequence of velocities executed during the iterative process. This discrepancy accumulates at each timestep, leading to inaccuracies.

\begin{figure}[!ht]
  \centering
  \includegraphics[width=8cm]{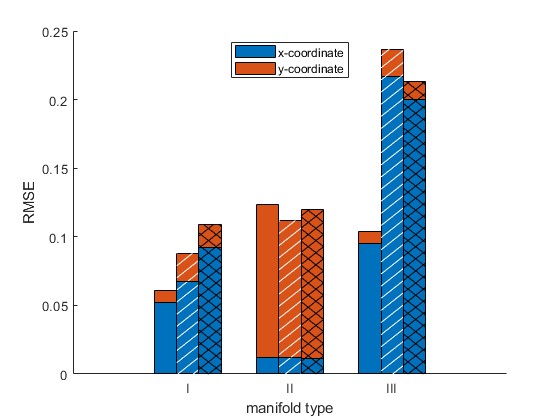}
  \caption{Comparison of RMSE for the data-streaming EDMD algorithm's prediction of follower trajectory for all types of manifold. The No Hatch Group uses leader-follower relative position only as a dictionary, the White Line Hatched Group uses both relative position and follower velocity as dictionaries, with follower's velocity data timestep $t = t-1$, and the Cross Line Hatched Group uses both relative position and follower velocity with data-streaming EDMD algorithm applied to follwer's velocity data.}
  \label{realcompare_bar}
\end{figure}

To demonstrate how these noise factors affect the data-streaming EDMD algorithm's estimation accuracy, Figure \ref{realcompare_bar} compares three different approaches across all three types of manifolds. The root mean square error (RMSE) between the estimated position and the ideal position serves as the metric for comparison. 

Contrary to Figure \ref{rmse_diction_el}, the results depicted in Figure \ref{realcompare_bar} indicate that, with the additional consideration of noise in the follower's velocity data, it is no longer a suitable dictionary for producing improved accuracy in the estimation of follower trajectory using data-streaming EDMD algorithm. The use of leader-follower relative position alone as the dictionary for the algorithm in estimating the follower trajectory emerges as the best choice for achieving optimal accuracy.

\section{Conclusion}

In this study, we introduce a follower trajectory estimation algorithm specifically tailored for maintaining a leader-follower column formation on an unknown, two-dimensional nonlinear manifold. Our approach utilize Koopman theory and Extended Dynamic Mode Decomposition (EDMD). The remarkable capability of the data-streaming EDMD algorithm in estimating follower trajectory has been demonstrated, as it significantly reduces the sensing range required for the follower to locate the ideal position provided by the leader. Through extensive simulations, we have determined that the leader-follower relative position measured in two-dimensional Euclidean metric serves as the most effective and practical choice for the data-streaming EDMD algorithm's dictionary. This local measurement can be effortlessly conducted by the follower using on-board sensors, rendering it highly feasible for real swarm robots. For future endeavors, we plan to implement actual swarm robots and conduct experiments to validate the efficacy of the approach proposed in this paper. Such empirical validation process holds the potential to establish a path toward practical efficient applications of swarm robotics in real-world scenarios. Specifically, it could advance capabilities in robot formation coordination within nonlinear environments.

\bibliographystyle{plain}        
\bibliography{references.bib}           

\begin{thebibliography}{10}

\bibitem{flockingattanasi}
Alessandro Attanasi, Andrea Cavagna, Lorenzo Del~Castello, Irene Giardina,
  Tomas~S. Grigera, Asja Jelić, Stefania Melillo, Leonardo Parisi, Oliver
  Pohl, Edward Shen, and Massimiliano Viale.
\newblock Information transfer and behavioural inertia in starling flocks.
\newblock {\em Nature Physics}, 10(9):691--696, 2014.

\bibitem{applychen}
Jianing Chen, Melvin Gauci, Wei Li, Andreas Kolling, and Roderich Groß.
\newblock Occlusion-based cooperative transport with a swarm of miniature
  mobile robots.
\newblock {\em IEEE Transactions on Robotics}, 31(2):307--321, 2015.

\bibitem{flockinggazi}
Veysel Gazi and Bariş Fidan.
\newblock Coordination and control of multi-agent dynamic systems: models and
  approaches, 2006.

\bibitem{applyGregory}
Jason~M. Gregory, Jonathan~R. Fink, Ethan Stump, Jeffrey~N. Twigg, John~G.
  Rogers, David~G. Baran, Nicholas Fung, and Stuart~H. Young.
\newblock Application of multi-robot systems to disaster-relief scenarios with
  limited communication.
\newblock In {\em International Symposium on Field and Service Robotics}, 2015.

\bibitem{Koopmanpaper}
B.~O. Koopman.
\newblock Hamiltonian systems and transformation in hilbert space.
\newblock {\em Proc Natl Acad Sci U S A}, 17(5):315--8, 1931.
\newblock 1091-6490 Koopman, B O Journal Article United States 1931/05/01 Proc
  Natl Acad Sci U S A. 1931 May;17(5):315-8. doi: 10.1073/pnas.17.5.315.

\bibitem{Mezic_koopmanMPC}
Milan Korda and Igor Mezić.
\newblock Linear predictors for nonlinear dynamical systems: Koopman operator
  meets model predictive control.
\newblock {\em Automatica}, 93:149--160, 2018.

\bibitem{EDMD_mezic}
Milan Korda and Igor Mezić.
\newblock On convergence of extended dynamic mode decomposition to the koopman
  operator.
\newblock {\em Journal of Nonlinear Science}, 28(2):687--710, 2018.

\bibitem{surveyLiu}
Yuanchang Liu and Richard Bucknall.
\newblock A survey of formation control and motion planning of multiple
  unmanned vehicles.
\newblock {\em Robotica}, 36(7):1019–1047, 2018.

\bibitem{flockingmateo}
David Mateo, Nikolaj Horsevad, Vahid Hassani, Mohammadreza Chamanbaz, and
  Roland Bouffanais.
\newblock Optimal network topology for responsive collective behavior.
\newblock {\em Science Advances}, 5(4):eaau0999, 2019.

\bibitem{Koopmanbook}
Alexandre Mauroy, Igor Mezic, and Yoshihiko~(Eds.) Susuki.
\newblock {\em The Koopman Operator in Systems and Control Concepts,
  Methodologies, and Applications: Concepts, Methodologies, and Applications}.
\newblock 01 2020.

\bibitem{applymcguire}
K.~N. McGuire, C.~De Wagter, K.~Tuyls, H.~J. Kappen, and G.~C. H.~E. de~Croon.
\newblock Minimal navigation solution for a swarm of tiny flying robots to
  explore an unknown environment.
\newblock {\em Science Robotics}, 4(35):eaaw9710, 2019.

\bibitem{applyMendonca}
Ricardo Mendonça, Mario~Monteiro Marques, Francisco Marques, André Lourenço,
  Eduardo Pinto, Pedro Santana, Fernando Coito, Victor Lobo, and José Barata.
\newblock A cooperative multi-robot team for the surveillance of shipwreck
  survivors at sea.
\newblock In {\em OCEANS 2016 MTS/IEEE Monterey}, pages 1--6, 2016.

\bibitem{Mezic_decomposition}
Igor Mezić.
\newblock Spectral properties of dynamical systems, model reduction and
  decompositions.
\newblock {\em Nonlinear Dynamics}, 41(1):309--325, 2005.

\bibitem{surveyOH}
Kwang-Kyo Oh, Myoung-Chul Park, and Hyo-Sung Ahn.
\newblock A survey of multi-agent formation control.
\newblock {\em Automatica}, 53:424--440, 2015.

\bibitem{flockingsaber}
Reza Olfati-Saber.
\newblock Flocking for multi-agent dynamic systems: algorithms and theory.
\newblock {\em IEEE Transactions on Automatic Control}, 51:401--420, 2006.

\bibitem{applyshapira}
Yaniv Shapira and Noa Agmon.
\newblock Path planning for optimizing survivability of multi-robot formation
  in adversarial environments.
\newblock In {\em 2015 IEEE/RSJ International Conference on Intelligent Robots
  and Systems (IROS)}, pages 4544--4549, 2015.

\bibitem{strogatz}
Steven~H. Strogatz.
\newblock {\em Nonlinear dynamics and Chaos : with applications to physics,
  biology, chemistry, and engineering}.
\newblock Studies in nonlinearity. Addison-Wesley Pub., Reading, Mass., 1994.

\bibitem{Vicsekflocking}
Tamás Vicsek and Anna Zafeiris.
\newblock Collective motion.
\newblock {\em Physics Reports}, 517(3):71--140, 2012.

\bibitem{Wang}
Yanran Wang and Takashi Hikihara.
\newblock Two-dimensional swarm formation in time-invariant external potential:
  Modeling, analysis, and control.
\newblock {\em Chaos: An Interdisciplinary Journal of Nonlinear Science},
  30(9):093145, 2020.

\bibitem{EDMD_rowley}
Matthew~O. Williams, Ioannis~G. Kevrekidis, and Clarence~W. Rowley.
\newblock A data–driven approximation of the koopman operator: Extending
  dynamic mode decomposition.
\newblock {\em Journal of Nonlinear Science}, 25(6):1307--1346, 2015.

\bibitem{applyzhang}
Siwei Zhang, Robert Pöhlmann, Thomas Wiedemann, Armin Dammann, Henk Wymeersch,
  and Peter~Adam Hoeher.
\newblock Self-aware swarm navigation in autonomous exploration missions.
\newblock {\em Proceedings of the IEEE}, 108(7):1168--1195, 2020.

\end{thebibliography}

\appendix
\section{Swarm formation construction}
\label{Swarm formation construction}
Let $U \subseteq \mathbb{R}^n$ be a non-empty open subset and $F: U \rightarrow \mathbb{R}$ a $C^{\infty}$. Let $\mathcal{M} \subseteq U \times \mathbb{R}$ be the
graph of $F$. The closed subset $\mathcal{M}$ in $U \times \mathbb{R}$ projects homeomorphically onto $U$ with inverse
$(x_1, \ldots, x_n) \mapsto (x_1, \ldots, x_n, F(x_1, \ldots, x_n))$
that is a smooth mapping from $U$ to $U \times \mathbb{R}$. $\mathcal{M}$ is a closed smooth submanifold of $U \times \mathbb{R}$. Using the standard Riemannian metric on $U \times \mathbb{R} \subseteq \mathbb{R}^{n+1}$, the induced metric $g$ on $\mathcal{M}$ at a point $p \in \mathcal{M}$ is
\begin{align}
 g(p) = \langle \partial_{q_i}|_p, \partial_{q_j}|_p\rangle_p \, dq_i(p) \otimes dq_j(p)
\end{align}
with coordinate chart $\{q_i\}$ on $\mathcal{M}$. Each $\partial_{q_i}|_p \in T_p\mathcal{M}$ can be represented as a linear combination of $\{\partial_{x_i}|_p\} \in T_p(\mathbb{R}^{n+1})$, given as
\begin{align}
 \partial_{q_i}|_p = \partial_{x_i}|_p + \partial_{x_i}f(p)\partial_{x_{n+1}}|_p.
\end{align}
Consider the aforementioned graph $\mathcal{M}$ as a $C^{\infty}$ Riemannian manifold. Given a curve, $C:[ a, b ] \longrightarrow \mathcal{M}$, a \textit{vector field} $X$ along $C$ is any section of the tangent bundle $T\mathcal{M}$ over $C$ ($X:[a,b]\longrightarrow T\mathcal{M}$, projection $\pi: T\mathcal{M} \longrightarrow \mathcal{M}$, such that $\pi \circ X = C$). If $\mathcal{M}$ is a smooth manifold, all vector field on the manifold are also smooth. We denote the collection of all smooth vector fields on manifold $\mathcal{M}$ as $\mathfrak{X}(\mathcal{M})$. For a Riemannian manifold $(\mathcal{M}, g)$, the \textit{Levi-Civita connection} $\nabla_g$ on $\mathcal{M}$ is the unique connection on $T\mathcal{M}$ that has both metric compatibility and torsion freeness. The Christoffel symbols of the second kind are the connection coefficients (in a local chart) of the Levi-Civita connection denoted as
\begin{align}
 \Gamma^a_{\,\,bc}=\frac{1}{2}g^{a d}(\partial_cg_{db}+\partial_bg_{dc}-\partial_dg_{bc}).
\end{align}
For a Riemannian manifold $(\mathcal{M}, g)$, a curve is called \textit{geodesic} with respect to the connection $\nabla_g$ if its acceleration it zero. That is a curve $\gamma$ where $\nabla_{\dot\gamma}\dot\gamma = 0$. A geodesic curve in n-dimensional Riemannian manifold can be expressed as a system of second order ordinary differential equations,
\begin{align}
 \frac{d^2\gamma^\lambda}{dt^2}+\Gamma^\lambda_{\,\,\mu \nu}\frac{d\gamma^\mu}{dt}\frac{d\gamma^\nu}{dt}=0.
 \label{geodesic formula}
\end{align}
All geodesics are the shortest path between any two points on the manifold.

We predefine the trajectory of the leader to be a geodesic curve on manifold. Base on (\ref{geodesic formula}), for a two-dimensional manifold $(\mathcal{M}, g)$, leader trajectory $l(\mathbf{x},t)$ can be expressed as a system of ordinary differential equations
\begin{align}
 \begin{split}
  \dot l_1 &= l_3,\\
  \dot l_2 &= l_4,\\
  \dot l_3 &= -\Gamma^x_{\,\,xx}(l_3)^2-2\Gamma^x_{\,\,xy}l_3l_4-\Gamma^x_{\,\,yy}(l_4)^2,\\
  \dot l_4 &=-\Gamma^y_{\,\,xx}(l_3)^2-2\Gamma^y_{\,\,xy}l_3l_4-\Gamma^y_{\,\,yy}(l_4)^2,
 \end{split}
 \label{head_curve_2d}
\end{align}
where basis $\{x,y\}$ are used in the index, and $\dot l$ are the first derivative with respect to time. Each Christoffel symbols depends entirely on the metric at a certain point in $\mathcal{M}$ in terms of basis $\{x,y\}$. Given initial conditions, $[x_1, x_2, \dot x_1, \dot x_2 ]$, (\ref{head_curve_2d}) is guaranteed to have a solution according to the Picard-Lindel$\rm{\ddot o}$f theorem. 

The distance of the edges of the ideal formation needs to be identical. For edges representation, a parallel vector field $V$ that is orthogonal to the leader trajectory $l(\mathbf{x}, t)$ is constructed as
\begin{align}
 \nabla_V V=0.
\end{align}
$V$ forms a family of geodesics where we require the initial position and velocity conditions $(u^j_0, v^j_0)$ for $j$th geodesic $\gamma_j \in V$  are $u^j_0=l(t_j)$, $v^j_0=v|\{g(v, \dot{l}(t_j))=0\}$, $j \in \mathbb{Z}$. 

The separation vector $s(t)$ connects a point $\gamma (t)$ on one geodesic to a point $\gamma(t)+s(t)$ on a nearby geodesic at the same time. For parallel vector field $V$, we can construct a separation vector field $S$ such that $s \in S$ are the separation vector discribed above. For the ideal formation, the followers' trajectories are the integral curves of $S$.

\section{Illustrations of algorithm effectiveness}
\label{Illustrations of algorithm effectiveness}

\begin{figure*}[h]
  \begin{center}
    \includegraphics[width=16cm]{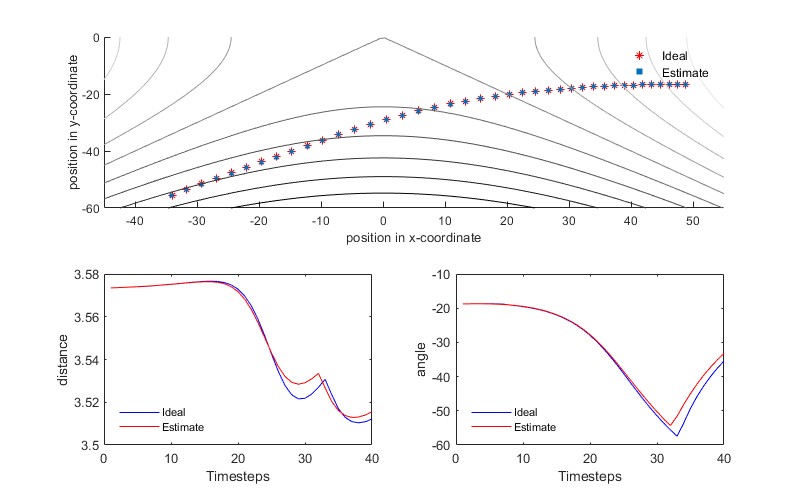}
  \end{center}
  \caption{Follower trajectory estimation by EDMD algorithm in type II manifold. The ideal trajectory points are in red, while estimated trajectory points in blue. Manifold show as the contour lines. }
  \label{graph_effective_hy}
\end{figure*}

In figure \ref{graph_effective_hy}, for the parameters influencing the follower trajectory, we selected the manifold as an hyperbolic paraboloid with the function $z = \frac{1}{30} (x^2-y^2$); the leader trajectory was defined as a geodesic curve of the manifold, starting with initial conditions $[x, y, \dot x, \dot y] = [50, -20, 1, 0]$.
  
\begin{figure*}[!h]
  \begin{center}
    \includegraphics[width=16cm]{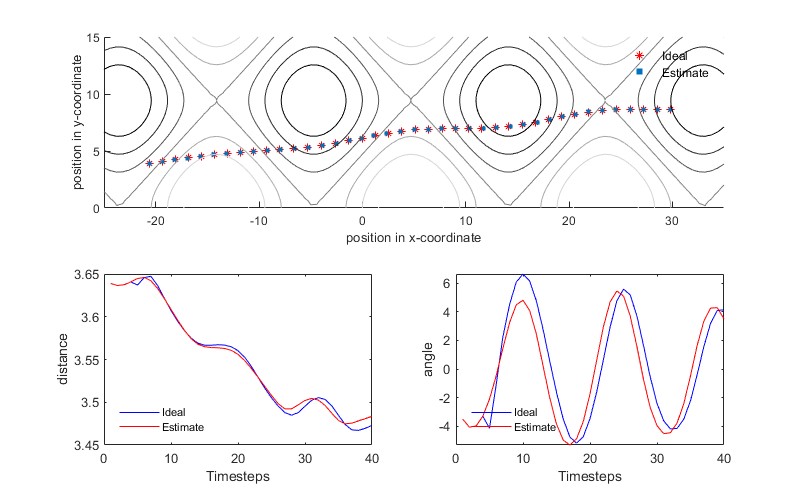}
  \end{center}
  \caption{Follower trajectory estimation by EDMD algorithm in type III manifold. The ideal trajectory points are in red, while estimated trajectory points in blue. Manifold show as the contour lines. }
  \label{graph_effective_trig}
\end{figure*}

In figure \ref{graph_effective_trig}, for the parameters influencing the follower trajectory, we selected manifold with the function $z = \sin(x/3)+\cos(y/3)$; the leader trajectory was defined as a geodesic curve of the manifold, starting with initial conditions $[x, y, \dot x, \dot y] = [30, 5, 1, 0]$.

\section{Dictionary for data-streaming EDMD algorithm}
Simulations conducted to compare how different
dictionaries affect the accuracy of the data-streaming EDMD
algorithm's estimation of ideal positions for manifold type II and type III. The specific dictionary choices are detailed in Table
1.
\label{dictionary for data-streaming EDMD algorithm}
\begin{figure}[h]
  \centering
  \begin{subfigure}[b]{0.49\textwidth}
      \centering
      \includegraphics[width=\textwidth]{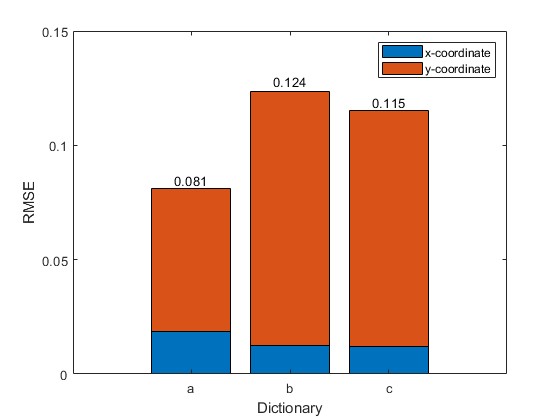}
      \caption{}
      \label{rmse_diction_hy}
  \end{subfigure}
  \hfill
  \begin{subfigure}[b]{0.49\textwidth}
      \centering
      \includegraphics[width=\textwidth]{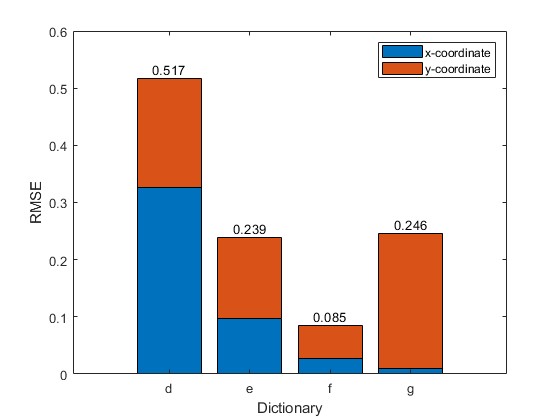}
      \caption{}
      \label{rmse_diction_withpos_hy}
  \end{subfigure}
  \hfill
  \caption{RMSE for various dictionary choices used for data-streaming EDMD algorithm using manifold type II parameters.}
\end{figure}

\begin{figure}[!h]
  \centering
  \begin{subfigure}[b]{0.49\textwidth}
      \centering
      \includegraphics[width=\textwidth]{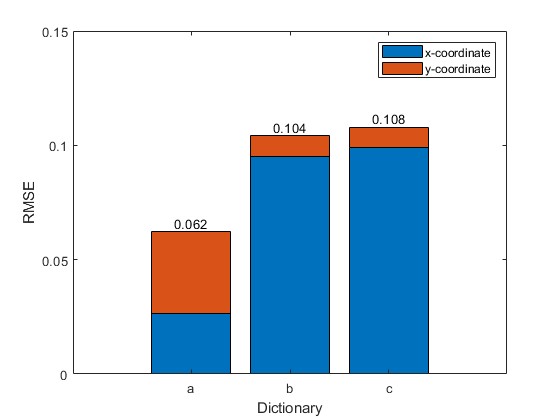}
      \caption{}
      \label{rmse_diction_trig}
  \end{subfigure}
  \hfill
  \begin{subfigure}[b]{0.49\textwidth}
      \centering
      \includegraphics[width=\textwidth]{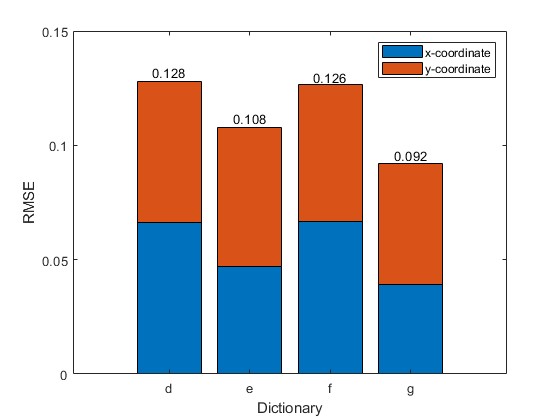}
      \caption{}
      \label{rmse_diction_withpos_trig}
  \end{subfigure}
  \hfill
  \caption{RMSE for various dictionary choices used for EDMD algorithm using manifold type II parameters.}
\end{figure}

\end{document}